\documentclass[10pt,twocolumn,letterpaper]{article}

\usepackage[pagenumbers]{cvpr}

\usepackage{times}
\usepackage{epsfig}
\usepackage{graphicx}
\usepackage{amsmath,amssymb}
\usepackage{booktabs}
\usepackage{multirow}
\usepackage{enumitem}
\usepackage{xspace}
\usepackage{siunitx}
\usepackage{subcaption}
\usepackage{array}
\usepackage{pifont}
\usepackage{graphicx}
\usepackage{tikz}
\usetikzlibrary{arrows.meta,calc}
\usepackage{cuted}
\usepackage{caption}
\usepackage{subcaption}
\usepackage{svg}
\usepackage{comment}
\usepackage{cuted}
\usepackage{caption}

\setlist[itemize]{leftmargin=*,topsep=2pt,itemsep=2pt,parsep=0pt}
\setlist[enumerate]{leftmargin=*,topsep=2pt,itemsep=2pt,parsep=0pt}

\newcolumntype{L}[1]{>{\raggedright\arraybackslash}p{#1}}
\newcolumntype{C}[1]{>{\centering\arraybackslash}p{#1}}
\newcolumntype{R}[1]{>{\raggedleft\arraybackslash}p{#1}}

\newcommand{\RoofDiTdot}{RoofDiT}
\newcommand{\RoofDiT}{\RoofDiTdot\space}

\definecolor{cvprblue}{rgb}{0.21,0.49,0.74}
\usepackage[
    pagebackref,
    breaklinks,
    colorlinks,
    allcolors=cvprblue
]{hyperref}

\def\confName{CVPR}
\def\confYear{2026}

\title{Diffusion Transformers for Roof Graph Synthesis and Reconstruction}

\author{
Daniel Panangian
\qquad
Ksenia Bittner
\\[4pt]
The Remote Sensing Technology Institute
\\
German Aerospace Center (DLR), Wessling, Germany
\\[3pt]
{\tt\small
\{daniel.panangian,ksenia.bittner\}@dlr.de}
}
\begin{document}

\maketitle

\begin{strip}
    \centering
    \includegraphics[width=0.95\textwidth]{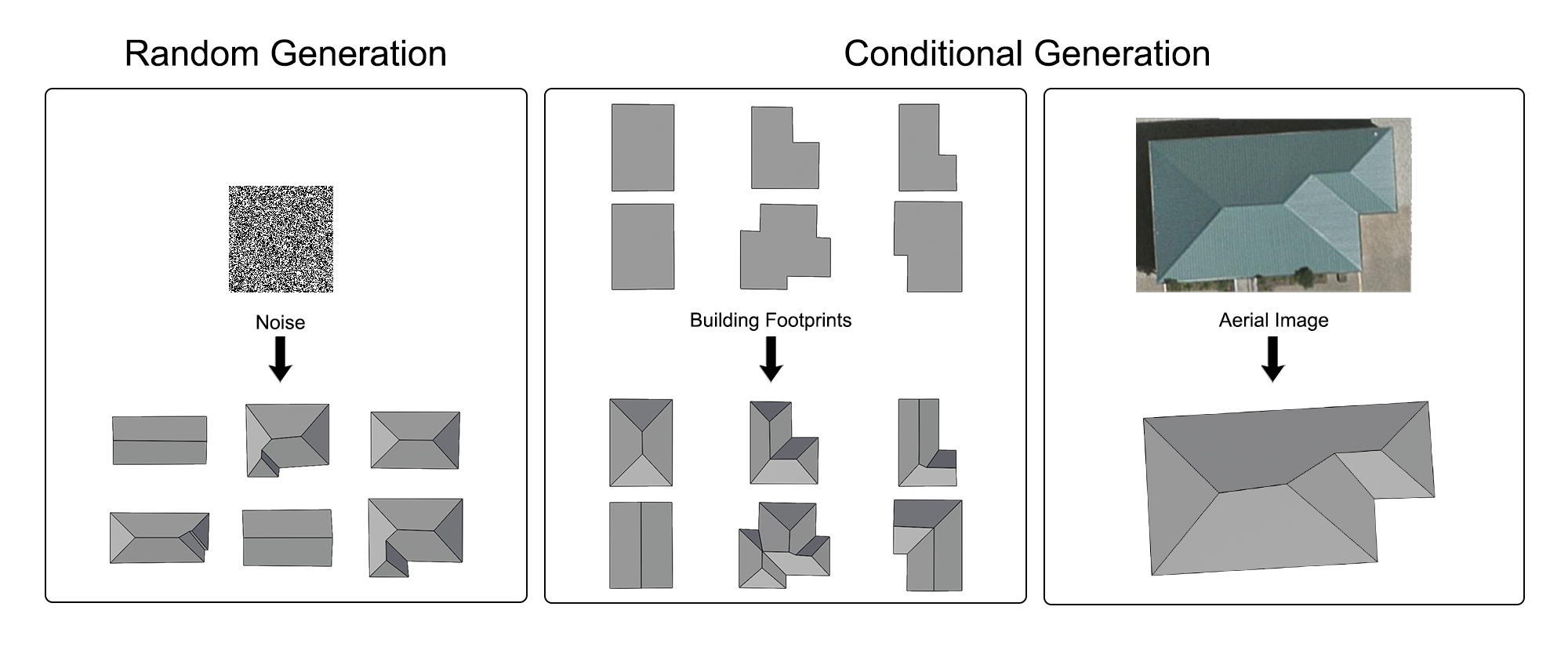}
    \captionof{figure}{
        Illustration of the three roof-graph generation tasks in our framework:
        (1) unconditional generation from noise,
        (2) conditional generation from building footprints, and
        (3) aerial-image-guided generation for reconstructing complete roof
        structures. The figure highlights our model's ability to sample
        plausible roof layouts from a learned prior and to perform constrained
        prediction using geometric and visual observations.
    }
    \label{fig:teaser}
\end{strip}

\begin{abstract}
We present \textsc{\RoofDiTdot}, a generative framework for 2d roof graph synthesis and reconstruction. The method represents roofs as vertex-edge graphs in top view and uses a two-stage pipeline. A diffusion transformer first generates roof graph vertices, and an edge prediction module then predicts graph connectivity. To improve geometric fidelity, RoofDiT incorporates relative geometry-aware attention, conditioning on footprint vertices and pretrained aerial image features, and a roof alignment regularizer that encourages horizontal, vertical, and diagonal structural patterns common in roof layouts. By varying the conditioning signal, the same model supports unconditional generation, footprint-conditioned generation, and image-guided reconstruction. Experiments on the benchmark show improved graph generation quality over a diffusion baseline in the unconditional setting, strong performance against a straight-skeleton prior in the footprint-conditioned setting, and competitive reconstruction results. Overall, our results demonstrate the potential of diffusion transformers for generative modeling of roof graphs.
\end{abstract}
\section{Introduction}
\label{sec:intro}

Residential roofs exhibit rich structural regularities  \cite{qian2021roofgan}. Their layouts are not arbitrary collections of edges and vertices, but follow recurring patterns in alignment, connectivity, and part composition. At the same time, roof structures remain diverse: buildings with similar overall shapes can realize different valid roof organizations. This combination of strong geometric constraint and nontrivial variation makes roof structure a compelling target for generative modeling. A useful model should capture both the regularities that make roof graphs structurally valid and the diversity that gives rise to multiple plausible configurations.
Recent years have seen substantial progress in generative modeling of structured visual and geometric data. Beyond image synthesis, generative models have been successfully applied to domains such as layouts, floorplans, and structured shape representations, where the objective is not only to produce realistic outputs but also to capture the underlying organization and constraints of the data. These advances suggest that generative models can serve as effective priors for domains in which validity depends on both local geometry and global structural consistency. Roof structure reconstruction remains challenging because the available observations often provide only partial evidence about the realized geometry. Many recent methods rely primarily on image-driven cues, such as roof lines, junctions, edges, texture, or learned feature maps from aerial imagery, to infer structural elements and assemble them into a roof layout \cite{zhao2022rsgnn, wang2025roofmapnet, chen2022heat}. These approaches have achieved strong performance, but they remain sensitive to ambiguity caused by occlusion, shadow, low resolution, and weak or missing visual boundaries. Other methods address this difficulty by incorporating stronger geometric priors, structural primitives, or topological constraints into the reconstruction process, improving robustness when image evidence alone is insufficient.

In this work, we investigate roof structure modeling through both unconditional and conditional generation. Given conditioning information such as building footprints or aerial imagery, the model is guided toward plausible structures consistent with the available evidence. We study three scenarios: random roof graph generation, footprint-conditioned generation, and image-guided reconstruction.~\Cref{fig:teaser} provides an overview of these tasks. Through this framework, we evaluate how a learned roof prior supports roof graph generation under different levels of conditioning.

Our contributions are summarized as follows:
\begin{itemize}
    \item We formulate roof structure modeling as a generative problem over 2D roof graphs and propose an approach for learning a prior over plausible roof structures.
    \item We adapt a diffusion-transformer model for roof graph generation, with support for conditioning on building footprints and aerial imagery.
    \item We benchmark the proposed method against roof generation and reconstruction baselines across unconditional, footprint-conditioned, and image-guided settings.
\end{itemize}

Overall, our results show that modeling a generative prior over roof graphs is useful beyond unconstrained sampling alone. It enables plausible roof generation without observations, controlled generation under footprint constraints, and improved reconstruction when aerial imagery is available.
\section{Related Work}
\label{sec:related}

\subsection{Generative Models for Layouts}
Generative modeling of structured spatial data has been widely studied for floorplans, layouts, and geometric design. Earlier methods often relied on raster or image-based representations, while more recent work has moved toward vector and graph formulations that better capture spatial structure. Representative examples include Graph2Plan~\cite{hu2020graph2plan}, House-GAN++~\cite{nauata2021houseganpp}, HouseDiffusion~\cite{shabani2023housediffusion}, and GSDiff~\cite{hu2024gsdiff}, as well as conditional layout generation methods~\cite{inoue2023layoutdm,hong2024cons2plan}. These works show that generative models can capture both structural regularity and output diversity under geometric constraints. Our work follows this line, but focuses on roof structures.

\begin{figure*}[t]
\begin{center}
\includegraphics[width=\linewidth]{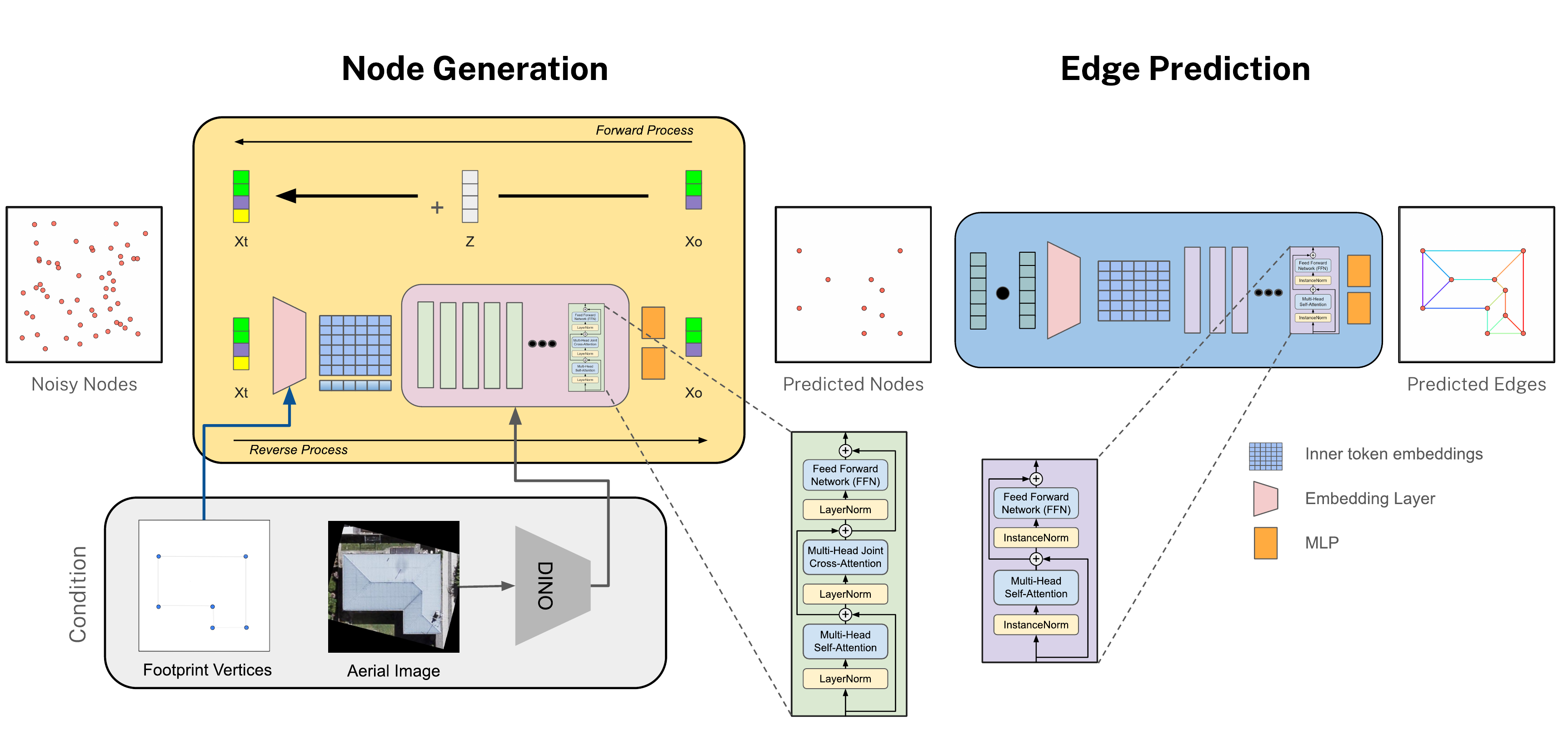}
\end{center}
\caption{Overall pipeline used in our method. Following GSDiff, the framework decomposes roof graph generation into two stages: node generation and edge prediction. In the first stage, a conditional diffusion transformer starts from noisy node states and iteratively denoises them to generate roof vertices, optionally conditioned on footprint vertices, pretrained aerial image features, or both. In the second stage, an edge prediction module takes the generated node set and infers pairwise connectivity to produce the final roof graph.}
\label{fig:architecture}
\end{figure*}

\subsection{Roof Structure Reconstruction from Aerial Imagery}
Roof reconstruction from aerial imagery has been studied as a structured prediction problem using geometric reasoning, graph inference, and learning-based methods~\cite{Li_2019_ICCV,zorzi2022polyworld}. Recent roof-specific approaches include RSGNN~\cite{zhao2022rsgnn}, HEAT~\cite{chen2022heat}, PolyRoof~\cite{polyroof2025}, and RoofMapNet~\cite{wang2025roofmapnet}, with segmentation-based variants also explored~\cite{xu2024multi}. These methods are closest to our application domain, but are mainly designed for deterministic reconstruction from image evidence. In contrast, our work studies roof prediction in a generative framework under different conditioning signals, including imagery and footprints.

\subsection{Roof Modeling and Generation}
Roofs have also been studied as objects of modeling and generation, especially in computer graphics. Classical methods use straight skeletons, procedural rules, or grammar-based constructions to derive roofs from building footprints~\cite{aichholzer1996skeleton,biedl2015weighted,eppstein1998raising,kelly2011interactive,muller2006procedural,buron2013gpuroof}. More recent work introduced roof-specific graph representations and optimization-based refinement for planar roof modeling~\cite{ren2021roofintuitive}, while learning-based generation remains relatively limited. Roof-GAN~\cite{qian2021roofgan} and related approaches~\cite{gtgan} highlight the value of learning structural roof priors. Our work differs by focusing on planar vertex-edge roof graphs as a generative 2D structural representation.

\section{Dataset}
\label{sec:data}

We use the residential roof dataset introduced by \citet{ren2021roofintuitive} as the basis for our experiments. The dataset contains annotated residential roofs together with corresponding aerial images, and is split into 1,926 training samples, 249 validation samples, and 223 test samples. We use this dataset because it provides a moderate-scale collection of residential roof structures with both geometric annotations and image observations, making it suitable for studying graph generation and image-guided prediction.

From the roof annotations, we derive 2D roof graphs in top view, where vertices correspond to roof junctions and edges correspond to roofline segments. Before training, we canonicalize the annotations to reduce unnecessary variation in orientation. Specifically, each roof is rotated so that the longest footprint edge aligns with a fixed reference axis. We then regularize edge directions by snapping nearly horizontal and vertical segments to the corresponding axis-aligned directions, which reduces small annotation inconsistencies while preserving the overall roof structure. For the footprint-conditioned setting, we additionally derive footprint vertices from the roof graph by taking the exterior boundary of the union of planar roof faces, and use these vertices as conditioning input.

\section{Approach}
\label{sec:method}

\subsection{Overview}
We build on the two-stage graph generation framework of GSDiff~\cite{hu2024gsdiff}, which generates nodes first and then predicts edges. We keep this pipeline across unconditional, footprint-conditioned, and image-guided settings, and modify the node generation stage with relative geometry-aware attention, multimodal conditioning, and a roof-specific alignment regularizer.

\subsection{Node Generation}
Let \(X \in \mathbb{R}^{N \times d}\) denote the node state of a roof graph, where \(N\) is the maximum number of nodes and \(d\) includes 2D coordinates and a validity/background flag. We model node generation as
\begin{equation}
p_{\theta}(X \mid c),
\end{equation}
where \(c\) is empty for unconditional generation and may include footprint vertices, image features, or both in conditioned settings. The node generator is implemented as a conditional diffusion transformer that denoises noisy node states into roof vertices.

\paragraph{Relative geometry-aware attention}
Self-attention over node tokens does not explicitly encode the geometric relations that are central to roof structure. For roof graphs, pairwise relations between vertices are highly informative because nearby vertices often belong to the same structural pattern, and directional relations help distinguish different roof layouts.

To incorporate this information, we augment attention with pairwise relative geometry features. For two nodes \(i\) and \(j\), we compute
\begin{equation}
\phi_{ij} = [\Delta x_{ij}, \Delta y_{ij}, r_{ij}, u_{x,ij}, u_{y,ij}],
\end{equation}
where \(\Delta x_{ij}\) and \(\Delta y_{ij}\) are coordinate offsets, \(r_{ij}\) is the Euclidean distance, and \((u_{x,ij}, u_{y,ij}) = (\Delta x_{ij}, \Delta y_{ij}) / r_{ij}\) denotes the normalized direction from node \(i\) to node \(j\). These features are mapped by a small multilayer perceptron to produce per-head attention biases, which are added to the attention logits. As a result, token interactions depend not only on learned token content but also on explicit geometric relations between roof vertices.
\paragraph{Conditioning design}
We use two complementary conditioning modalities for node generation: building footprints and aerial imagery. For footprint-conditioned generation, footprint vertices are encoded as 2D geometric tokens in the same representation space as the generated roof nodes and concatenated to the node sequence, so the transformer processes roof nodes and footprint vertices jointly through self-attention. This enables reasoning over interactions between the known outer building geometry and the internal roof structure, while relative geometry-aware attention also provides pairwise geometric biases between roof nodes and footprint vertices. For image-guided reconstruction, features extracted by a pretrained DINOv2~\cite{oquab2023dinov2} encoder are projected into a feature memory with positional encoding and injected through cross-attention. The two conditioning sources can also be combined, with footprint vertices entering the joint node stream and image features provided through a separate branch.

\paragraph{Roof alignment}
In addition to the geometry-aware node generator, we use a lightweight roof alignment regularizer during training. Roof layouts often contain not only horizontal and vertical alignments but also diagonal support lines. We therefore define the regularizer over four canonical undirected line families:
\begin{equation}
\left\{
\begin{aligned}
x &= \mathrm{const},\\
y &= \mathrm{const},\\
x + y &= \mathrm{const},\\
x - y &= \mathrm{const}.
\end{aligned}
\right.
\end{equation}
The first two correspond to vertical and horizontal alignment, while the latter two capture the two diagonal families that frequently occur in roof layouts.

For each valid node, we measure its smallest support-line distance to any other valid node under these four families. Let \(d^x_{ij}\), \(d^y_{ij}\), \(d^{+}_{ij}\), and \(d^{-}_{ij}\) denote the corresponding pairwise distances between nodes \(i\) and \(j\). We define the best support-line distance for node \(i\) as
\begin{equation}
m_i = \min_{j \neq i} \min \left( d^x_{ij}, d^y_{ij}, d^{+}_{ij}, d^{-}_{ij} \right).
\end{equation}
These distances are then converted into penalties with time-dependent weighting over diffusion steps. This regularizer does not enforce roof topology directly, but provides an additional geometric bias that is better matched to common roof layouts and complements the node generator during training.

The final node generation objective combines the diffusion loss with the roof alignment term:
\begin{equation}
\mathcal{L}_{\text{node}}
=
\mathcal{L}_{\text{diff}}
+
\lambda_{\text{align}} \mathcal{L}_{\text{roof-align}},
\end{equation}
where \(\lambda_{\text{align}}\) controls the contribution of the alignment regularizer.

\subsection{Edge Prediction}
After node generation, the second stage predicts graph connectivity from the generated node set. Concretely, given the generated roof vertices, the edge prediction module evaluates candidate node pairs and predicts whether an edge should connect them. This stage corresponds to modeling the connectivity term conditioned on the generated nodes, and yields the final roof graph. We keep the edge prediction stage unchanged from the underlying two-stage graph pipeline. 
\section{Experiments}
\label{sec:experiments}

\subsection{Experimental Setup}
We use the train/validation/test split described in~\cref{sec:data}. We evaluate three settings: unconditional generation, footprint-conditioned generation, and image-guided reconstruction. In all cases, \RoofDiT uses two-stage prediction with node generation followed by edge prediction. For footprint-conditioned generation, the model is conditioned on the building footprint; for image-guided reconstruction, aerial image features are additionally used. In the generative settings, inference is stochastic, and for footprint-conditioned generation we sample five outputs per test footprint and report averaged and best-of-5 metrics where appropriate. For image-guided reconstruction, inference is deterministic and uses a single prediction per input. All models are trained with AdamW for 100{,}000 steps with batch size 128, learning rate \(10^{-4}\), and a cosine diffusion schedule of 1000 steps. For image-guided reconstruction, aerial images are encoded using a frozen pretrained DINOv2 ViT-S/14 backbone.

\subsection{Compared Methods}
We compare \RoofDiT with \textbf{GSDiff} for unconditional generation, \textbf{straight skeleton} for footprint-conditioned generation, and \textbf{HEAT} plus \textbf{RoofMapNet} for image-guided reconstruction. HEAT is retrained on the same split as \RoofDiTdot, while RoofMapNet is evaluated using its released pretrained model. Since these baselines use pixel-aligned roof annotations whereas \RoofDiT uses canonicalized graph representations, the image-guided results should be interpreted as reference comparisons rather than strictly controlled benchmarks.
\subsection{Evaluation Metrics}
We use different protocols for the three settings. For unconditional generation, the test split is too small for stable distribution-level evaluation, so we generate 1000 roof graphs and compare them against repeated balanced subsets of the test set using FID and KID, following GSDiff. We also report Roof Valid Rate, Planar Rate, and Duplicate Rate@3.

For footprint-conditioned generation, five samples are generated per test footprint. Since a footprint may correspond to multiple plausible roofs, we report best-of-5 node, edge, and face metrics. Geometry metrics such as Valid Rate and Planar Rate are averaged over all samples. For image-guided reconstruction, all metrics are computed on a single prediction.
A roof graph is considered valid if it is non-empty, planar, free of dangling degree-one nodes, and contains at least one cycle. Roof Valid Rate and Planar Rate are defined as
\[
\mathrm{RoofValidRate}
=
\frac{1}{N}\sum_{i=1}^{N}\mathbf{1}\{G_i \text{ is valid}\}.
\]
\[
\mathrm{PlanarRate}
=
\frac{1}{N}\sum_{i=1}^{N}\mathbf{1}\{G_i \text{ has no edge crossings}\}.
\]

Duplicate Rate@3 measures node redundancy after projecting node coordinates to a \(256 \times 256\) grid:
\[
\mathrm{DupRate@3}
=
\frac{1}{N}\sum_{i=1}^{N}
\frac{1}{|V_i|}
\sum_{v_j \in V_i}
\mathbf{1}
\left\{
\min_{k \neq j}\|p_j - p_k\|_2 < 3
\right\}.
\]

Node evaluation uses Hungarian matching under a spatial threshold, from which we compute precision, recall, F1, and node count MAE. Edge evaluation counts an edge as correct when both endpoints are matched and connectivity is preserved. Face evaluation follows the HEAT region-based protocol, reporting precision, recall, F1, matched IoU, and face count error. We additionally report the number of components, crossings, and dangling nodes as geometry metrics, where lower values indicate better graph quality. Unless otherwise stated, node and edge metrics are reported at a 5-pixel threshold.
\section{Results}
\label{sec:results}

\begin{table}[!t]
\centering
\caption{Performance on unconditional roof graph generation}
\label{tab:uncond_generation}
\begin{tabular}{lcc}
\toprule
Metric & GSDiff & RoofDiT \\
\midrule
Roof Valid Rate $\uparrow$ & 0.941 & \textbf{0.943} \\
Planar Rate $\uparrow$ & 0.952 & \textbf{0.958} \\
Duplicate Rate@3 $\downarrow$ & 0.004 & \textbf{0.002} \\
FID$\downarrow$ & 53.954 & \textbf{45.329} \\
KID$\downarrow$ & 31.210 & \textbf{23.442} \\
\bottomrule
\end{tabular}
\end{table}

\begin{table}[h]
\centering
\caption{Performance on footprint-conditioned roof graph generation. SS denotes the straight skeleton baseline.}
\label{tab:footprint_conditioned_generation}
\setlength{\tabcolsep}{4pt}
\renewcommand{\arraystretch}{0.98}
\footnotesize
\begin{tabular}{llcc}
\toprule
 & Metric & SS & RoofDiT \\
\midrule
\multirow{2}{*}{Node}
& Count MAE $\downarrow$ & 2.008 & \textbf{1.002} \\
& F1@5 $\uparrow$ & \textbf{0.849} & 0.705 \\
\midrule
Edge
& F1 $\uparrow$ & 0.960 & \textbf{0.981} \\
\midrule
\multirow{4}{*}{Face}
& Count Err. $\downarrow$ & 1.430 & \textbf{0.444} \\
& F1 $\uparrow$ & \textbf{0.840} & 0.798 \\
& F1$_{\text{best-of-}K}$ $\uparrow$ & 0.840 & \textbf{0.886} \\
& Matched IoU $\uparrow$ & \textbf{0.850} & 0.697 \\
\midrule
\multirow{3}{*}{Geometry}
& Planar Rate $\uparrow$ & \textbf{1.000} & 0.873 \\
& Valid Rate $\uparrow$ & \textbf{1.000} & 0.841 \\
& Valid Rate$_{\text{best-of-}K}$ $\uparrow$ & \textbf{1.000} & 0.930 \\
\bottomrule
\end{tabular}
\end{table}

\begin{figure*}[t]
    \centering
    \begin{subfigure}[t]{0.48\textwidth}
        \centering
        \includegraphics[width=\textwidth]{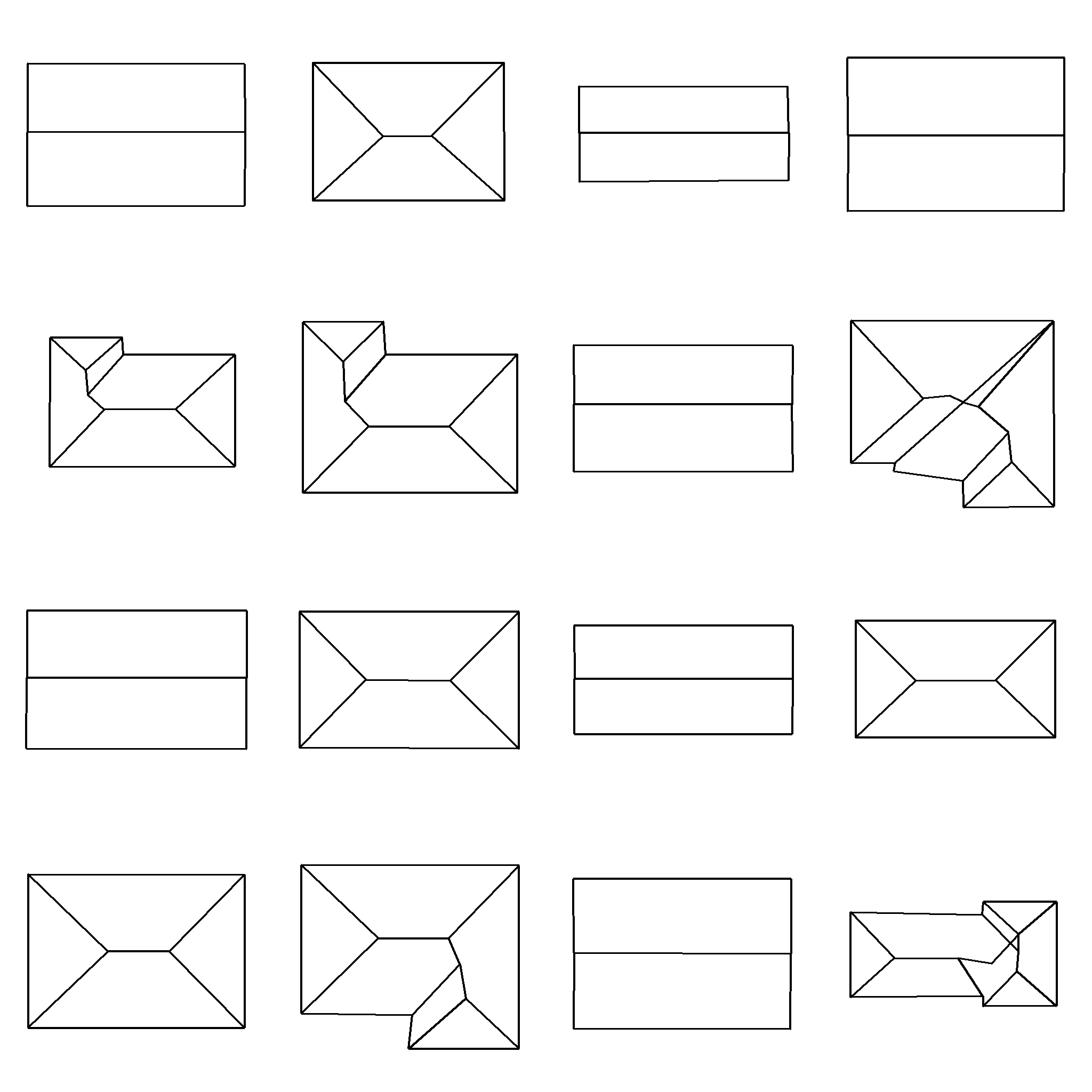}
        \caption{GSDiff}
        \label{fig:gsdiff_grid}
    \end{subfigure}
    \hfill
    \begin{subfigure}[t]{0.48\textwidth}
        \centering
        \includegraphics[width=\textwidth]{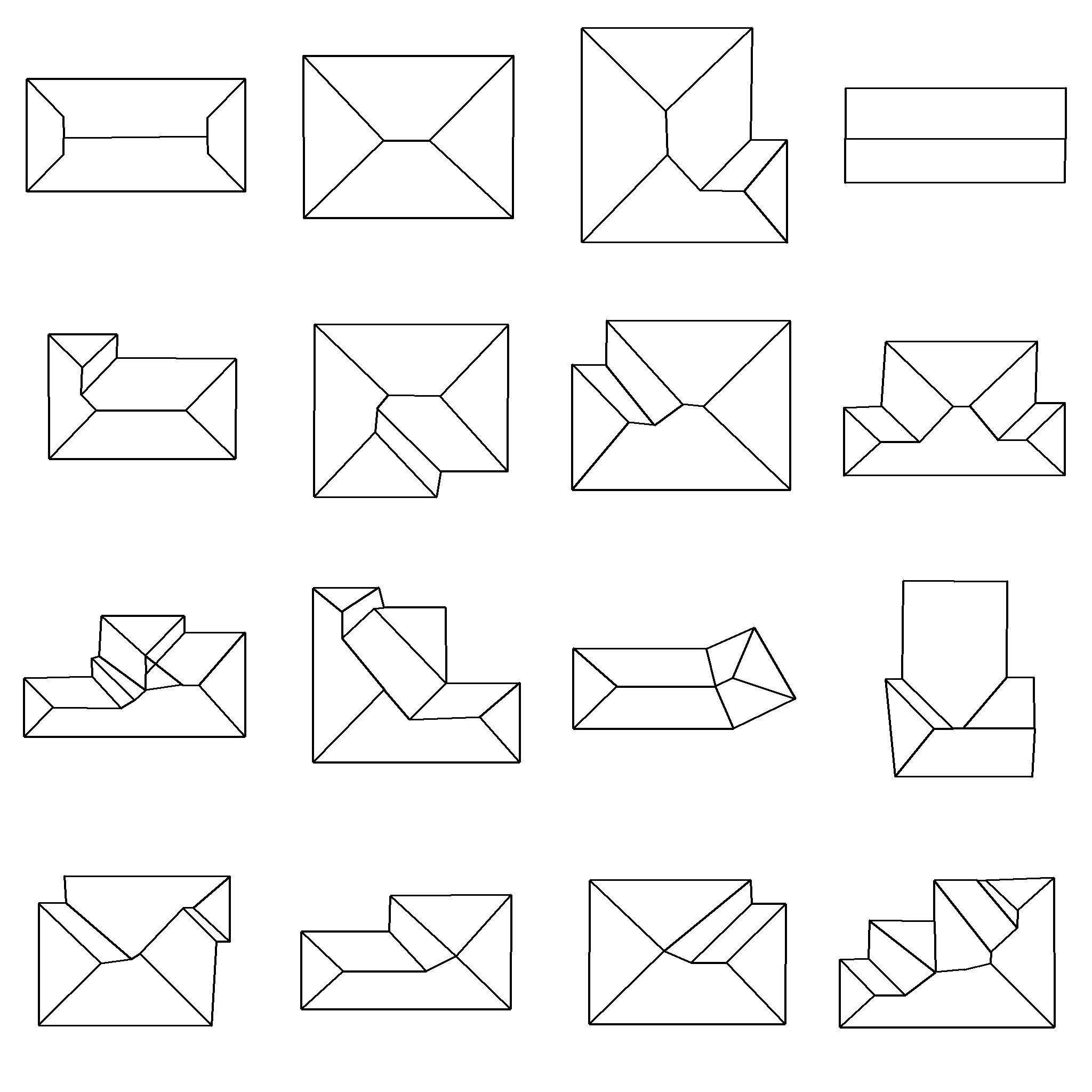}
        \caption{RoofDiT}
        \label{fig:roofdiff_grid}
    \end{subfigure}
    \caption{Qualitative comparison of unconditional roof graph generation. Samples produced by GSDiff and RoofDiT are shown side by side.}
    \label{fig:gsdiff_vs_roofdiff}
\end{figure*}

\begin{figure*}[!t]
\centering
\setlength{\tabcolsep}{0pt}

\begin{subfigure}{0.20\textwidth}
    \centering
    \includegraphics[width=0.96\linewidth]{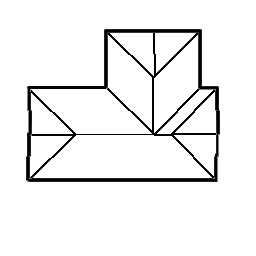}
\end{subfigure}\hfill
\begin{subfigure}{0.40\textwidth}
    \centering
    \includegraphics[width=0.475\linewidth]{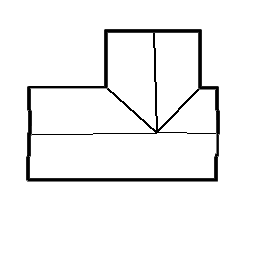}
    \hfill
    \includegraphics[width=0.475\linewidth]{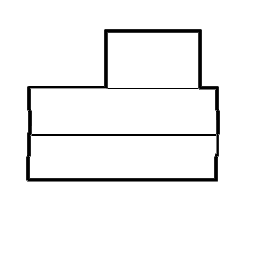}
\end{subfigure}\hfill
\begin{subfigure}{0.20\textwidth}
    \centering
    \includegraphics[width=0.96\linewidth]{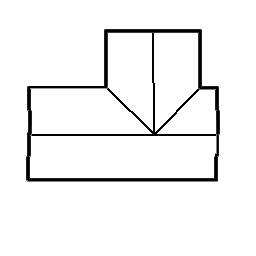}
\end{subfigure}\\[-7mm]

\begin{subfigure}{0.20\textwidth}
    \centering
    \includegraphics[width=0.96\linewidth]{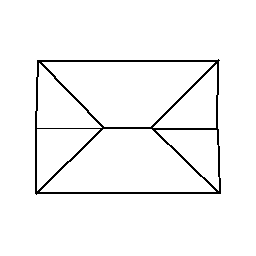}
\end{subfigure}\hfill
\begin{subfigure}{0.40\textwidth}
    \centering
    \includegraphics[width=0.475\linewidth]{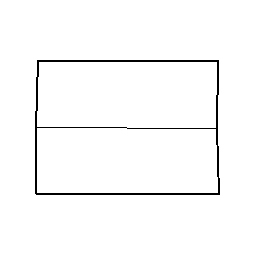}
    \hfill
    \includegraphics[width=0.475\linewidth]{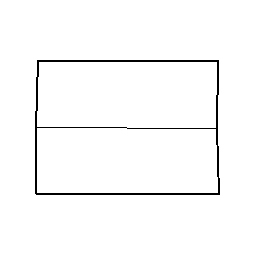}
\end{subfigure}\hfill
\begin{subfigure}{0.20\textwidth}
    \centering
    \includegraphics[width=0.96\linewidth]{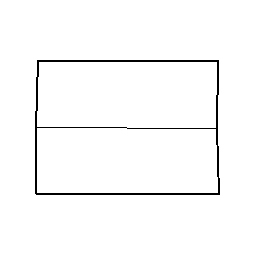}
\end{subfigure}\\[-7mm]

\begin{subfigure}{0.20\textwidth}
    \centering
    \includegraphics[width=0.96\linewidth]{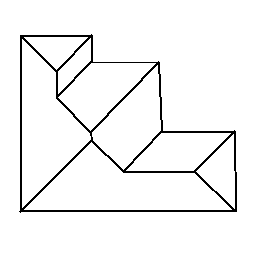}
    \caption{Straight Skeleton}
\end{subfigure}\hfill
\begin{subfigure}{0.40\textwidth}
    \centering
    \includegraphics[width=0.475\linewidth]{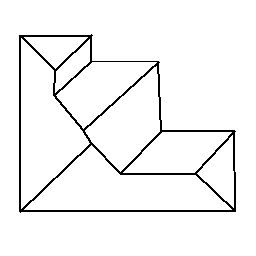}
    \hfill
    \includegraphics[width=0.475\linewidth]{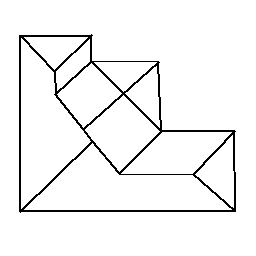}
    \caption{RoofDiT}
\end{subfigure}\hfill
\begin{subfigure}{0.20\textwidth}
    \centering
    \includegraphics[width=0.96\linewidth]{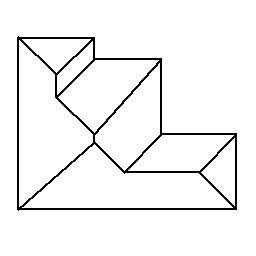}
    \caption{GT}
\end{subfigure}

\caption{Qualitative comparison for footprint-conditioned roof graph generation}
\label{fig:roof_qual_row}
\end{figure*}

\subsection{Unconditional Generation}

Table~\ref{tab:uncond_generation} compares unconditional roof graph generation between RoofDiT and GSDiff. RoofDiT improves all reported metrics, with slightly higher roof valid and planar rates, lower duplicate rate, and clearly lower FID and KID. The small gaps in validity-related metrics indicate that both methods usually generate valid, planar graphs, while the larger gains in FID and KID suggest that RoofDiT better matches the distribution of realistic roof graphs. These trends are also reflected in \cref{fig:gsdiff_vs_roofdiff}. GSDiff often produces simpler and more repetitive layouts dominated by basic symmetric ridge patterns, whereas RoofDiT generates a broader range of roof graphs, including both simple and more complex asymmetric configurations. This greater diversity makes the samples closer to the variability observed in real roof graphs, although some outputs are also more intricate. Overall, RoofDiT improves both distributional realism and diversity.

\subsection{Footprint-Conditioned Generation}

\Cref{tab:footprint_conditioned_generation} compares footprint-conditioned generation between the straight skeleton baseline and RoofDiT. RoofDiT achieves substantially lower node count MAE and face count error, and improves edge F1 from 0.960 to 0.981. Under best-of-$K$ evaluation, it further improves face F1 from 0.798 to 0.886 and valid rate from 0.841 to 0.930. These results indicate that RoofDiT better recovers the overall amount of roof structure and benefits from sampling multiple plausible candidates for the same footprint. At the same time, the straight skeleton baseline remains stronger on node F1@5, face F1, matched IoU, planar rate, and valid rate, reflecting its handcrafted geometric prior and deterministic structural regularity. Overall, the results show a tradeoff between the stronger geometric guarantees of straight skeleton and the greater flexibility of RoofDiT. The qualitative examples in~\cref{fig:roof_qual_row} illustrate this tradeoff. Straight skeleton can only return a fixed procedural interpretation of the footprint, often overpredicting nodes and faces through unnecessary internal partitions. In the shown examples, its outputs respect the footprint but appear mechanically derived and less faithful to the target roof organization. RoofDiT, in contrast, often produces samples that are closer to the ground truth in ridge layout and face decomposition, while also generating diverse yet plausible alternatives for the same footprint. This is important because a building footprint does not uniquely determine a single roof topology.
\subsection{Image-guided Reconstruction}

\Cref{tab:image_conditioned_node_edge_face_geometry} compares image-guided roof graph reconstruction across node, edge, face, and geometry metrics. RoofMapNet performs substantially worse than the other methods across all metric groups. HEAT achieves the strongest overall performance, with the best node precision, recall, and F1, the strongest face metrics, and the best geometry scores, including the highest valid and planar rates and the fewest crossings. For RoofDiT, adding footprint conditioning improves performance over using aerial imagery alone across all metric groups: node F1 increases from 42.6 to 70.3, edge F1 from 78.4 to 98.4, face F1 from 63.5 to 82.4, and node count MAE decreases from 2.23 to 0.90. Valid rate also increases from 87.4 to 88.3, and planar rate from 90.1 to 90.6. When both aerial imagery and building footprints are provided, RoofDiT achieves the strongest edge precision, recall, and F1 among the compared methods. Because this variant uses additional structural input, this is not a strictly like-for-like comparison with image-only baselines. Overall, HEAT remains the strongest image-only baseline, while RoofDiT provides a practical footprint-augmented alternative when building footprints are available.

\begin{table*}[!t]
\caption{Comparison of image-conditioned roof graph reconstruction methods grouped by node, edge, face, and geometry metrics. Node metrics are reported at a 5-pixel matching threshold. RoofDiT$_{\text{aerial}}$ uses aerial imagery only, while RoofDiT$_{\text{aerial+footprint}}$ additionally uses the building footprint as input.}
\centering
\footnotesize
\resizebox{\textwidth}{!}{%
\begin{tabular}{lccccccccccccccccc}
\toprule
& \multicolumn{4}{c}{Node} & \multicolumn{3}{c}{Edge} & \multicolumn{5}{c}{Face} & \multicolumn{5}{c}{Geometry} \\
\cmidrule(lr){2-5} \cmidrule(lr){6-8} \cmidrule(lr){9-13} \cmidrule(lr){14-18}
Method & Prec & Recall & F1 & Count MAE$\downarrow$ & Prec & Recall & F1 & Prec & Recall & F1 & Matched IoU & Count Err.$\downarrow$ & Valid & Planar & \#Comp$\downarrow$ & Crossings$\downarrow$ & Dangling$\downarrow$ \\
\midrule
RoofMapNet
& 27.9 & 40.8 & 32.7 & 7.40
& 43.6 & 45.9 & 44.5
& 25.5 & 27.8 & 26.1 & 25.9 & 0.30
& 20.2 & 38.1 & 5.12 & 3.22 & 1.51 \\

HEAT
& \textbf{93.3} & \textbf{93.0} & \textbf{93.0} & \textbf{0.30}
& 93.9 & 94.2 & 94.0
& \textbf{91.6} & \textbf{89.9} & \textbf{90.2} & \textbf{86.2} & \textbf{0.11}
& \textbf{90.6} & \textbf{92.8} & \textbf{1.00} & \textbf{0.11} & 0.09 \\

\midrule
RoofDiT$_{\text{aerial}}$
& 42.7 & 43.9 & 42.6 & 2.23
& 78.1 & 80.5 & 78.4
& 63.7 & 66.2 & 63.5 & 60.4 & 1.12
& 87.4 & 90.1 & \textbf{1.00} & 0.13 & \textbf{0.04} \\

RoofDiT$_{\text{aerial+footprint}}$
& 70.5 & 70.9 & 70.3 & 0.90
& \textbf{98.5} & \textbf{98.5} & \textbf{98.4}
& 81.9 & 83.8 & 82.4 & 71.8 & 0.48
& 88.3 & 90.6 & \textbf{1.00} & 0.19 & \textbf{0.04} \\
\bottomrule
\end{tabular}%
}
\label{tab:image_conditioned_node_edge_face_geometry}
\end{table*}
\begin{figure*}[!t]
\centering

\begin{tabular}{ccccc}
Input & HEAT & RoofMapNet & RoofDiT & GT \\[1mm]

\includegraphics[width=0.15\textwidth]{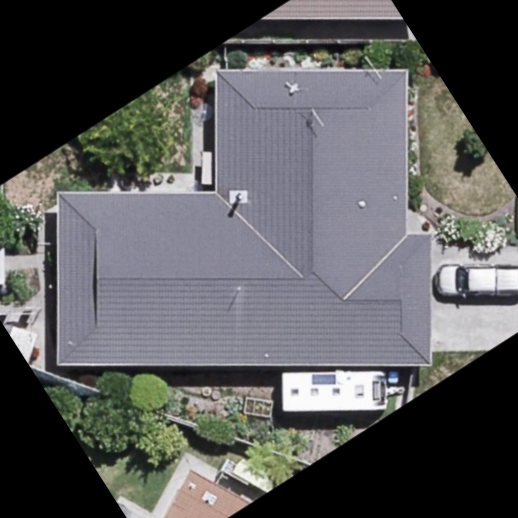} &
\includegraphics[width=0.15\textwidth]{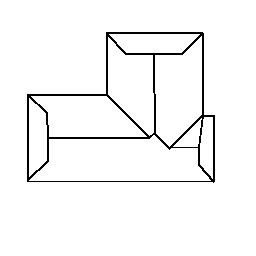} &
\includegraphics[width=0.15\textwidth]{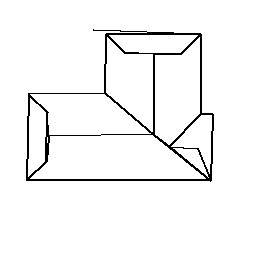} &
\includegraphics[width=0.15\textwidth]{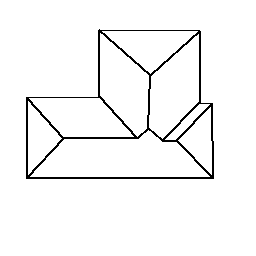} &
\includegraphics[width=0.15\textwidth]{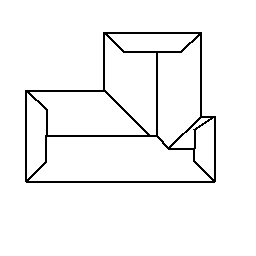} \\[-1mm]

\includegraphics[width=0.15\textwidth]{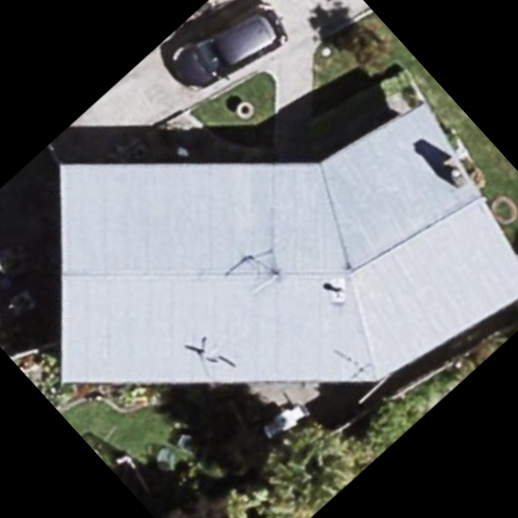} &
\includegraphics[width=0.15\textwidth]{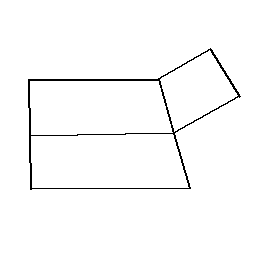} &
\includegraphics[width=0.15\textwidth]{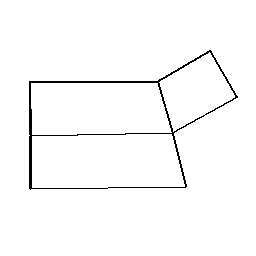} &
\includegraphics[width=0.15\textwidth]{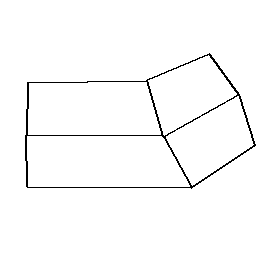} &
\includegraphics[width=0.15\textwidth]{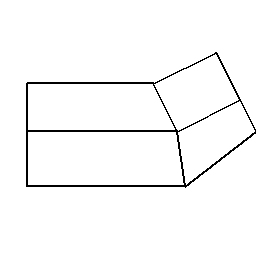} \\[-1mm]

\includegraphics[width=0.15\textwidth]{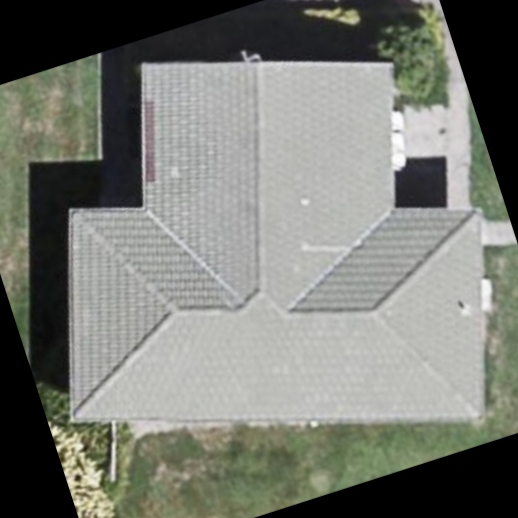} &
\includegraphics[width=0.15\textwidth]{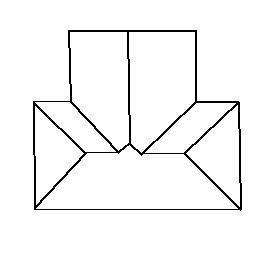} &
\includegraphics[width=0.15\textwidth]{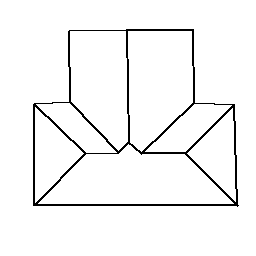} &
\includegraphics[width=0.15\textwidth]{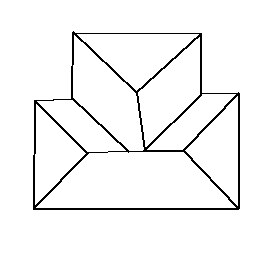} &
\includegraphics[width=0.15\textwidth]{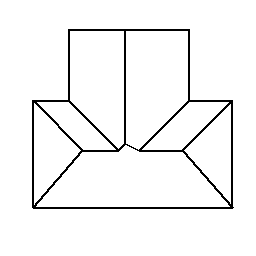} \\[-1mm]

\includegraphics[width=0.15\textwidth]{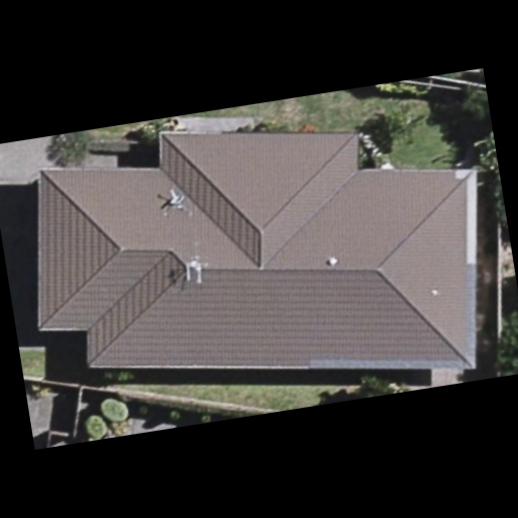} &
\includegraphics[width=0.15\textwidth]{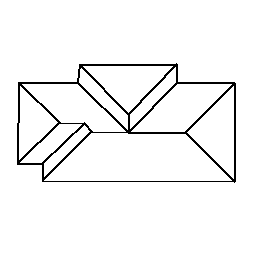} &
\includegraphics[width=0.15\textwidth]{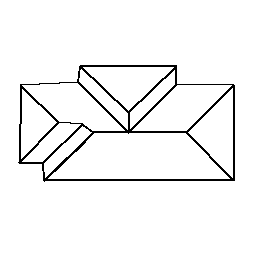} &
\includegraphics[width=0.15\textwidth]{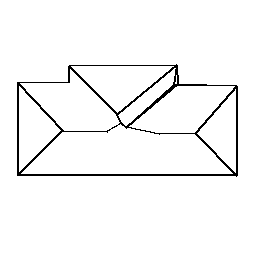} &
\includegraphics[width=0.15\textwidth]{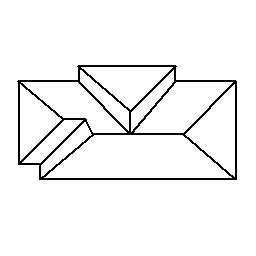} \\

\end{tabular}

\caption{Qualitative comparison of image-guided roof graph reconstruction. RoofDiT is conditioned on the input image only, while HEAT and RoofMapNet are shown for comparison against the ground-truth roof graph.}
\label{fig:qual_6col}
\end{figure*}

The qualitative examples in~\cref{fig:qual_6col} complement the quantitative results. HEAT remains the strongest reconstruction baseline in Table~\ref{tab:image_conditioned_node_edge_face_geometry}, while RoofDiT is particularly strong on edge recovery and often produces regular, globally coherent roof graphs. In the first row, RoofDiT reconstructs a plausible roof graph with a clean overall structure, although HEAT is slightly closer to the ground truth in some local details. In the second row, RoofDiT performs best among the compared methods: while the other methods fail to recover the simple target structure, RoofDiT captures the main topology more faithfully. This suggests that RoofDiT can be more robust in some ambiguous cases even when aggregate metrics favor HEAT. In the third row, RoofDiT again recovers a plausible roof family and the main global layout, showing that it can handle more articulated roofs when the image evidence is informative. In the fourth row, however, RoofDiT misses a face that is visible in the input and present in the ground truth. Overall, RoofDiT can produce strong and regular predictions, but not always the closest match to the ground truth.
\section{Discussion}
\label{sec:discussion}

The results suggest that roof structure can be modeled effectively as a learned prior over planar roof graphs. Across the three settings, RoofDiT does more than generate plausible samples: it also captures structural regularities that remain useful under conditioning. This is especially visible in the unconditional setting, where the model produces valid and diverse roof graphs that better match the target distribution than the baseline. More broadly, this indicates that the model learns meaningful regularities of roof topology rather than relying only on conditioning signals at inference time. The footprint-conditioned setting provides a particularly informative view of this prior. The comparison with the Straight Skeleton baseline highlights an important tradeoff. Geometric constructions offer strong guarantees of validity and regularity, while a learned generative model can represent a broader range of plausible roof organizations. The gains under best-of-$K$ evaluation reinforce this point. They suggest that RoofDiT often places good solutions within its sample distribution, even when a single draw does not always select the closest one. This points less to a limitation in representational capacity than to a limitation in consistency or sample selection.The image-guided reconstruction results show similar patterns. Although HEAT still performs better overall in precise agreement with ground truth, RoofDiT remains competitive and is particularly strong in edge recovery. This is notable because edge structure carries much of the roof topology. The clear improvement from adding footprint information further suggests that the generative formulation is most useful when visual evidence alone is insufficient and additional structural cues help constrain the solution space.

Several limitations remain. The current model operates on planar roof graphs and does not directly recover full 3D geometry, roof heights, or watertight building models. In addition, validity is encouraged through learning rather than guaranteed by construction, which leaves geometric baselines stronger in strict validity and deterministic consistency. At the same time, the overall results indicate a promising direction: the model already captures useful structural knowledge, and improvements in training scale, sampling strategy, and geometric constraints may reduce the current weaknesses while preserving the flexibility of generative prediction.

\section{Conclusion}
\label{sec:conclusion}

This work shows that generative roof graph modeling is a promising approach for both roof graph generation and evidence-guided roof reconstruction. Across unconditional and footprint-conditioned generation, RoofDiT demonstrates that a diffusion-based model can learn a meaningful prior over plausible planar roof graphs. In image-guided reconstruction, the method remains competitive with strong baselines and can effectively exploit footprint information when it is available.

Overall, the results suggest that learned generative modeling provides a useful complement to deterministic geometric methods. While geometric baselines still offer stronger guarantees in validity and consistency, the generative approach is better suited to representing multiple plausible roof graphs when the available evidence does not uniquely determine the solution. This makes it a promising foundation for future work on more reliable, structurally constrained, and geometrically richer roof reconstruction models.

{
    \small
    \bibliographystyle{ieeenat_fullname}
    \bibliography{main}

@String(ICCV= {Int. Conf. Comput. Vis.})

@String(ICCV  = {ICCV})

@article{ren2021roofintuitive,
  author    = {Ren, Jing and Zhang, Biao and Wu, Bojian and Huang, Jianqiang and Fan, Lubin and Ovsjanikov, Maks and Wonka, Peter},
  title     = {Intuitive and Efficient Roof Modeling for Reconstruction and Synthesis},
  journal   = {ACM Transactions on Graphics},
  year      = {2021},
  volume    = {40},
  number    = {6},
  articleno = {249},
  numpages  = {17},
  month     = dec,
  issn      = {0730-0301},
  publisher = {Association for Computing Machinery},
  address   = {New York, NY, USA},
  doi       = {10.1145/3478513.3480494},
  url       = {https://doi.org/10.1145/3478513.3480494}
}

@article{zhao2022rsgnn,
  author   = {Zhao, Wufan and Persello, Claudio and Stein, Alfred},
  title    = {Extracting Planar Roof Structures from Very High Resolution Images Using Graph Neural Networks},
  journal  = {ISPRS Journal of Photogrammetry and Remote Sensing},
  year     = {2022},
  volume   = {187},
  pages    = {34--45},
  issn     = {0924-2716},
  doi      = {10.1016/j.isprsjprs.2022.02.022},
  url      = {https://www.sciencedirect.com/science/article/pii/S092427162200065X}
}

@article{wang2025roofmapnet,
  author   = {Wang, Jiaqi and Chen, Guanzhou and Zhang, Xiaodong and Wang, Tong and Tan, Xiaoliang and Yang, Qingyuan and Zhou, Wenlin and Zhu, Kun},
  title    = {RoofMapNet: Utilizing Geometric Primitives for Depicting Planar Building Roof Structure from High-Resolution Remote Sensing Imagery},
  journal  = {International Journal of Applied Earth Observation and Geoinformation},
  year     = {2025},
  volume   = {141},
  pages    = {104630},
  issn     = {1569-8432},
  doi      = {10.1016/j.jag.2025.104630},
  url      = {https://www.sciencedirect.com/science/article/pii/S1569843225002778}
}

@inproceedings{chen2022heat,
  author    = {Chen, Jiacheng and Qian, Yiming and Furukawa, Yasutaka},
  title     = {HEAT: Holistic Edge Attention Transformer for Structured Reconstruction},
  booktitle = {Proceedings of the IEEE/CVF Conference on Computer Vision and Pattern Recognition},
  year      = {2022},
  pages     = {3856--3865},
  month     = jun,
  publisher = {IEEE Computer Society},
  address   = {Los Alamitos, CA, USA},
  doi       = {10.1109/CVPR52688.2022.00384},
  url       = {https://doi.ieeecomputersociety.org/10.1109/CVPR52688.2022.00384}
}

@inproceedings{qian2021roofgan,
  author    = {Qian, Yiming and Zhang, Hao and Furukawa, Yasutaka},
  title     = {Roof-GAN: Learning to Generate Roof Geometry and Relations for Residential Houses},
  booktitle = {Proceedings of the IEEE/CVF Conference on Computer Vision and Pattern Recognition},
  year      = {2021},
  pages     = {2795--2804},
  url       = {https://api.semanticscholar.org/CorpusID:229298053}
}

@article{hu2020graph2plan,
  author    = {Hu, Ruizhen and Huang, Zeyu and Tang, Yuhan and Van Kaick, Oliver and Zhang, Hao and Huang, Hui},
  title     = {Graph2Plan: Learning Floorplan Generation from Layout Graphs},
  journal   = {ACM Transactions on Graphics},
  year      = {2020},
  volume    = {39},
  number    = {4},
  articleno = {118},
  numpages  = {14},
  month     = aug,
  issn      = {0730-0301},
  publisher = {Association for Computing Machinery},
  address   = {New York, NY, USA},
  doi       = {10.1145/3386569.3392391},
  url       = {https://doi.org/10.1145/3386569.3392391}
}

@inproceedings{nauata2021houseganpp,
  author    = {Nauata, Nelson and Hosseini, Sepidehsadat and Chang, Kai-Hung and Chu, Hang and Cheng, Chin-Yi and Furukawa, Yasutaka},
  title     = {House-GAN++: Generative Adversarial Layout Refinement Network towards Intelligent Computational Agent for Professional Architects},
  booktitle = {Proceedings of the IEEE/CVF Conference on Computer Vision and Pattern Recognition},
  year      = {2021},
  pages     = {13627--13636},
  doi       = {10.1109/CVPR46437.2021.01342}
}

@inproceedings{shabani2023housediffusion,
  author    = {Shabani, Mohammad Amin and Hosseini, Sepidehsadat and Furukawa, Yasutaka},
  title     = {HouseDiffusion: Vector Floorplan Generation via a Diffusion Model with Discrete and Continuous Denoising},
  booktitle = {Proceedings of the IEEE/CVF Conference on Computer Vision and Pattern Recognition},
  year      = {2023},
  pages     = {5466--5475}
}

@article{hu2024gsdiff,
  author  = {Hu, Sizhe and Wu, Wenming and Wang, Yuntao and Xu, Benzhu and Zheng, Liping},
  title   = {GSDiff: Synthesizing Vector Floorplans via Geometry-Enhanced Structural Graph Generation},
  journal = {arXiv preprint arXiv:2408.16258},
  year    = {2024}
}

@inproceedings{inoue2023layoutdm,
  author    = {Inoue, Naoto and Kikuchi, Kotaro and Simo-Serra, Edgar and Otani, Mayu and Yamaguchi, Kota},
  title     = {LayoutDM: Discrete Diffusion Model for Controllable Layout Generation},
  booktitle = {Proceedings of the IEEE/CVF Conference on Computer Vision and Pattern Recognition},
  year      = {2023},
  pages     = {10167--10176},
  doi       = {10.1109/CVPR52729.2023.00980}
}

@inproceedings{hong2024cons2plan,
  author    = {Hong, Shibo and Zhang, Xuhong and Du, Tianyu and Cheng, Sheng and Wang, Xun and Yin, Jianwei},
  title     = {Cons2Plan: Vector Floorplan Generation from Various Conditions via a Learning Framework Based on Conditional Diffusion Models},
  booktitle = {Proceedings of the 32nd ACM International Conference on Multimedia},
  year      = {2024},
  pages     = {3248--3256},
  numpages  = {9},
  publisher = {Association for Computing Machinery},
  address   = {New York, NY, USA},
  isbn      = {9798400706868},
  doi       = {10.1145/3664647.3680681},
  url       = {https://doi.org/10.1145/3664647.3680681},
  location  = {Melbourne VIC, Australia},
  series    = {MM '24}
}

@article{xu2024multi,
  title={Multi-branch convolutional neural network in building polygonization using remote sensing images},
  author={Xu, Yajin and Schuegraf, Philipp and Bittner, Ksenia},
  journal={PFG--Journal of Photogrammetry, Remote Sensing and Geoinformation Science},
  volume={93},
  number={1},
  pages={79--100},
  year={2025},
  publisher={Springer}
}

@inproceedings{zorzi2022polyworld,
  title={Polyworld: Polygonal building extraction with graph neural networks in satellite images},
  author={Zorzi, Stefano and Bazrafkan, Shabab and Habenschuss, Stefan and Fraundorfer, Friedrich},
  booktitle={Proceedings of the IEEE/CVF Conference on Computer Vision and Pattern Recognition},
  pages={1848--1857},
  year={2022}
}

@InProceedings{Li_2019_ICCV,
author = {Li, Zuoyue and Wegner, Jan Dirk and Lucchi, Aurelien},
title = {Topological Map Extraction From Overhead Images},
booktitle = {Proceedings of the IEEE/CVF International Conference on Computer Vision (ICCV)},
month = {October},
year = {2019}
}

@inproceedings{polyroof2025,
  author       = {Chaikal Amrullah and
                  Daniel Panangian and
                  Ksenia Bittner},
  title        = {PolyRoof: Precision Roof Polygonization in Urban Residential Building
                  with Graph Neural Networks},
  booktitle    = {Joint Urban Remote Sensing Event, {JURSE} 2025, Tunis, Tunisia, May
                  5-7, 2025},
  pages        = {1--4},
  publisher    = {{IEEE}},
  year         = {2025},
  url          = {https://doi.org/10.1109/JURSE60372.2025.11075990},
  doi          = {10.1109/JURSE60372.2025.11075990},
  bibsource    = {dblp computer science bibliography, https://dblp.org}
}

@ARTICLE{gtgan,
  author={Tang, Hao and Shao, Ling and Sebe, Nicu and Van Gool, Luc},
  journal={IEEE Transactions on Pattern Analysis and Machine Intelligence}, 
  title={Graph Transformer GANs With Graph Masked Modeling for Architectural Layout Generation}, 
  year={2024},
  volume={46},
  number={6},
  pages={4298-4313},
  doi={10.1109/TPAMI.2024.3355248}}

@inproceedings{muller2006procedural,
  title = {Procedural modeling of buildings},
  author = {Pascal M{\"u}ller and Peter Wonka and Simon Haegler and Andreas Ulmer and {Van Gool}, Luc},
  year = {2006},
  month = dec,
  day = {1},
  doi = {10.1145/1179352.1141931},
  language = {English (US)},
  isbn = {1595933646},
  series = {ACM SIGGRAPH 2006 Papers, SIGGRAPH '06},
  pages = {614--623},
  booktitle = {ACM SIGGRAPH 2006 Papers, SIGGRAPH '06},
  note = {ACM SIGGRAPH 2006 Papers, SIGGRAPH '06 ; Conference date: 30-07-2006 Through 03-08-2006}
}

@article{kelly2011interactive,
  title = {Interactive architectural modeling with procedural extrusions},
  author = {Tom Kelly and Peter Wonka},
  year = {2011},
  month = apr,
  doi = {10.1145/1944846.1944854},
  language = {English (US)},
  volume = {30},
  journal = {ACM Transactions on Graphics},
  issn = {0730-0301},
  publisher = {Association for Computing Machinery (ACM)},
  number = {2}
}

@inproceedings{eppstein1998raising,
  author = {Eppstein, David and Erickson, Jeff},
  title = {Raising roofs, crashing cycles, and playing pool: applications of a data structure for finding pairwise interactions},
  year = {1998},
  isbn = {0897919734},
  publisher = {Association for Computing Machinery},
  address = {New York, NY, USA},
  url = {https://doi.org/10.1145/276884.276891},
  doi = {10.1145/276884.276891},
  booktitle = {Proceedings of the Fourteenth Annual Symposium on Computational Geometry},
  pages = {58--67},
  numpages = {10},
  location = {Minneapolis, Minnesota, USA},
  series = {SCG '98}
}

@inproceedings{buron2013gpuroof,
  booktitle = {Eurographics 2013 - Short Papers},
  editor = {M.- A. Otaduy and O. Sorkine},
  title = {GPU Roof Grammars},
  author = {Buron, Cyprien and Marvie, Jean-Eudes and Gautron, Pascal},
  year = {2013},
  publisher = {The Eurographics Association},
  issn = {1017-4656},
  doi = {10.2312/conf/EG2013/short/085-088}
}

@article{biedl2015weighted,
  title = {Weighted straight skeletons in the plane},
  journal = {Computational Geometry},
  volume = {48},
  number = {2},
  pages = {120--133},
  year = {2015},
  issn = {0925-7721},
  doi = {10.1016/j.comgeo.2014.08.006},
  url = {https://www.sciencedirect.com/science/article/pii/S0925772114000807},
  author = {Therese Biedl and Martin Held and Stefan Huber and Dominik Kaaser and Peter Palfrader}
}

@inproceedings{aichholzer1996skeleton,
  author = {Aichholzer, Oswin and Aurenhammer, Franz},
  editor = {Cai, Jin-Yi and Wong, Chak Kuen},
  title = {Straight skeletons for general polygonal figures in the plane},
  booktitle = {Computing and Combinatorics},
  year = {1996},
  publisher = {Springer Berlin Heidelberg},
  address = {Berlin, Heidelberg},
  pages = {117--126},
  isbn = {978-3-540-68461-9}
}

@misc{oquab2023dinov2,
  title={DINOv2: Learning Robust Visual Features without Supervision},
  author={Oquab, Maxime and Darcet, Timothée and Moutakanni, Theo and Vo, Huy V. and Szafraniec, Marc and Khalidov, Vasil and Fernandez, Pierre and Haziza, Daniel and Massa, Francisco and El-Nouby, Alaaeldin and Howes, Russell and Huang, Po-Yao and Xu, Hu and Sharma, Vasu and Li, Shang-Wen and Galuba, Wojciech and Rabbat, Mike and Assran, Mido and Ballas, Nicolas and Synnaeve, Gabriel and Misra, Ishan and Jegou, Herve and Mairal, Julien and Labatut, Patrick and Joulin, Armand and Bojanowski, Piotr},
  journal={arXiv:2304.07193},
  year={2023}
}
}


\end{document}